\documentclass[10pt]{IEEEtran}
\usepackage[
  margin=2cm,
  includefoot,
  footskip=30pt,
]{geometry}
\usepackage{latexsym}
\usepackage{algpseudocode}
\usepackage{algorithm}
\usepackage{graphicx}
\usepackage[numbers,sort&compress]{natbib}
\usepackage{url}
\usepackage{float}

\newfloat{axiomatization}{thp}{lop}
\floatname{axiomatization}{Class Definition}

\begin{document}
\pagenumbering{gobble}
\title{Ontology of Card Sleights}
\author{Aaron Sterling\\
        Department of Computer Science\\
        Iowa State University\\
        Ames, Iowa, USA\\
        sterling@iastate.edu}
\maketitle
\begin{abstract}
We present a machine-readable movement writing for sleight-of-hand moves with cards---a ``Labanotation of card magic.'' This scheme of movement writing contains 440 categories of motion, and appears to taxonomize all card sleights that have appeared in over 1500 publications. The movement writing is axiomatized in $\mathcal{SROIQ}$(D) Description Logic, and collected formally as an Ontology of Card Sleights, a computational ontology that extends the Basic Formal Ontology and the Information Artifact Ontology. The Ontology of Card Sleights is implemented in OWL DL, a Description Logic fragment of the Web Ontology Language. While ontologies have historically been used to classify at a less granular level, the algorithmic nature of card tricks allows us to transcribe a performer's actions step by step. We conclude by discussing design criteria we have used to ensure the ontology can be accessed and modified with a simple click-and-drag interface. This may allow database searches and performance transcriptions by users with card magic knowledge, but no ontology background.
\end{abstract}
\pagenumbering{arabic}
\section{Introduction}
\subsection{Overview}
In 1928, dance theorist Rudolph Laban invented Labanotation, a way to write down dance moves, much as the staff notation of sharps and flats provides a way to write down music~\cite{labanotation}. While there have been other methods of movement writing, before and since, Labanotation has been the one most used by academics and movement professionals. Nevertheless, the use and teaching of Labanotation has declined as it has become easier to videorecord a dance performance~\cite{dance-legacy}. Computer scientists have tried to build systems of machine-readable movement representation since the 1970s~\cite{badler1979digital}. Just as there are advantages of a musical score over a musical recording---easier to search, easier to compare with other pieces of music---so too, there are advantages of a dance score over a dance video. Nevertheless, it is hard to escape the fact that a highly trained expert is required to produce Labanotation, while ``anyone'' can pull out their phone and record a video. It remains an open problem to create a movement writing that (1) captures a nontrivial aspect of human motion, (2) is amenable to complex database queries, and (3) can be written and read by someone who is not a movement writing expert.

We believe that we have solved this open problem within the movement domain of card magic. Card tricks have an explicitly algorithmic character: there is a well-defined starting state, the steps from one state to the next are mathematical transformations, and there is a well-defined ending state. While the ``magic-ness'' of the trick often depends on complex social context, if we strip away all script and social interaction, we are left with a ``recipe'' of both visible and secret moves that is called the \emph{method}. In this paper, we present a way to record card trick methods using OWL DL, a Description Logic fragment of the Web Ontology Language. Description Logics are often used in knowledge representation systems, because they are decidable and have a formal semantics, so reasoning programs can obtain a large body of inferences from a relatively small base of knowledge~\cite{description-logic-primer}. The Web Ontology Language provides a means by which movement can be recorded unambiguously and shared~\cite{owl-overview}. Finally, we built the Ontology of Card Sleights by extending the Basic Formal Ontology~\cite{arp2015building}, which also provides the framework for medical and scientific ontologies whose user bases have science background but little or no ontology background (see for example~\cite{hastings2012chebi}). Therefore, we believe our ontology of movement can be used by people with magic background but no ontology or computer science background.
\subsection{Computer science background}
See~\cite{dance-notation} for a survey of different methods of dance notation. Many researchers have applied dance notation to computer science, and vice-versa; see~\cite{robot-motion} for several recent examples. The work that is closest to our own is that of El Raheb and Ioannidis, who are building a Labanotation-inspired ontology to preserve traditional Greek folk dances~\cite{dance-ontology, el2013dance}. El Raheb and Ioannidis report on DanceOWL, a dance ontology built in OWL DL. They define about 300 different classes of body movement and they provide a way to record the passage of time. We consider their work excellent, but we found that we could not use their approach, for reasons we now explain.

El Raheb and Ioannidis work with an ``everything-equal'' approach, where each version of a particular dance is just as valid as any other. The researchers are finding the dances ``in nature,'' instead of setting a particular performance as the canonical example of all dances of type $X$. Indeed, they use a particular flag to denote ``this is a score,'' and that is the only thing that separates a score from the recording of a particular performance; they intentionally require that a score be given no greater importance than a performance. By contrast, the world of magic is full of controversies over proper credit, and who published what when is extremely important. For our movement domain of interest, we needed a clear way to determine whether, ``Performances $P$ and $Q$ are both performances of the same trick $T$.'' To evaluate that, one must determine not the distance between $P$ and $Q$, but the distance between $P$ and $T$ and the distance between $Q$ and $T$, where $T$ is the official version of the trick. Moreover, the ``everything-equal'' approach does not allow for one of the main arguments movement professionals make about the benefits of Labanotation over video: with a written score, one can separate the intent of the original creator from the interpretation of a choreographer, and from dance steps that might have been improvised in a single performance~\cite{dance-legacy}.

It is logically nontrivial to separate the instructions of how to perform something from a transcription of a performance. Consider, for example, what it would take to express formally that the English version, the French version and the Japanese version of a novel, including different editions with different typographical errors, are all the same book. We wanted the ability to be able to say exactly this, about magic tricks performed by different performers in different countries. That was what led us to ground our work within the Basic Formal Ontology (hereafter BFO). To continue the example of the novel, in BFO, the ``true'' version of the novel is a \texttt{generically dependent continuant}, while a hard copy of the English-language second edition is a \texttt{material entity} that is a subclass of \texttt{independent continuant}. Working within such a framework creates overhead, so it is fair to ask why one can't simply build an ontology that is a naive tagging system, much like the tags one might put on a YouTube video.

The creation of BFO was, in part, motivated by the prevalence of ontological limitations like the following. Imagine a database with three ontological classes: \texttt{Person}, \texttt{Shift Worker} and \texttt{Manager}. The universe is that of individual persons, and they can either be shift workers or managers. While this works fine within the context of a payroll database, we know that in real life, there was a time when person $P$ was neither a shift worker nor a manager for that company. The equation of personhood with employment in the company prevents that ontology from being applied more generally. We would like a movement writing that can be used to transcribe an entire magic show, or to transcribe every trick in a magic book. To that end, we need to be able to track individual playing cards beyond their context in a particular trick. BFO allows us to say that person $P$ is a material entity, who plays the \emph{role of Shift Worker} or the \emph{role of Manager} at time $t$. A role is a \texttt{specifically dependent continuant}---a property that depends upon a specific material entity at a specific point in time. Person $P$ may change roles over time while still remaining person $P$. This approach allows us to track playing cards (and other props) across performance contexts, regardless of what roles the props might play in a given trick.

More abstractly, we also needed a way to categorize the many card tricks that have been published but never performed, and a way to handle the many card tricks and card sleights (secret moves) that are minor variations of other tricks and sleights. The Information Artifact Ontology~\cite{iao}, an extension of BFO, provided us the tools to handle this. Roughly speaking, there are four types of entities that appear in the Ontology of Card Sleights: (1) objects (like Playing Cards), (2) qualities (properties of objects, like whether a Playing Card is face up or face down), (3) Card Actions (processes in which objects participate, like turning over a Playing Card), and (4) instructional entities (directions for how to do something, like, ``Turn over the top card''). We obtain (1)-(3) from BFO, and (4) from the Information Artifact Ontology (hereafter IAO).
\subsection{Magic background}
There have been many attempts to categorize theatrical magic by plot or effect (for both a classic and more recent example see~\cite{magic-mirror, neo-magic}). Focusing on card magic, Giobbi in 2006 published a classification of ``Basic Effects of Card Magic''~\cite{giobbi-basic-effects}. We started our work in this area also, attempting to build an ontology of card tricks by theme or effect~\cite{ontology-of-card-magic}. However, we soon realized that we needed a lower-level language first, so a database could answer questions like, ``Does this card trick already exist in the literature?'' There are many different methods one can employ to achieve the same effect, so this paper describes our attempt to create an ontology of method.

We owe a great debt to card expert Denis Behr and a project he leads called the Conjuring Archive~\cite{conjuring-archive}. The Conjuring Archive is an online resource that has categorized thousands of magic tricks, especially card tricks, by effect, and has also categorized thousands of card sleights, that have appeared in over 1500 publications. We began our work with the categorization of card sleights in the Conjuring Archive, and kept about three-fourths of those categories. About a fourth of the categories we had to eliminate or change, because the Conjuring Archive employs a naive tagging system, and some categories did not fit with a BFO-structured ontology. We provide examples of our changes in Section~\ref{section:ocs}. We now present some technical details of the Ontology of Card Sleights.

%
\section{Ontology of card sleights}
\subsection{Technical preliminaries}
The current version of the Ontology of Card Sleights is just over 18,000 lines of OWL DL code\footnote{The Ontology of Card Sleights is available for download at \url{purl.org/net/ontologyofcardsleights}. It is best viewed in Prot{\'e}g{\'e} 5, whose installation instructions can be found at \url{https://protegewiki.stanford.edu/wiki/Install_Protege5}.} We designed it using the ontology editor Prot{\'e}g{\'e} 5~\cite{protege}. Our ontology heavily uses classes from BFO 2.0, and extends two classes from the IAO. The most important classes in the ontology are:
\begin{enumerate}
\item \texttt{Card Action}: This class is the primary work product of the ontology. It extends the BFO class \texttt{process} and contains 440 subclasses of actions with cards. Most of these actions are sleights, but there are also what we term \texttt{Straight Card Action}, actions that are exactly what they appear to be.
\item \texttt{Theatrical Magic Object}: This class extends the BFO class \texttt{object}, and its subclasses include various props, the most often used of which are \texttt{Playing Card} and \texttt{Table}.
\item \texttt{Card Magic Object Aggregate}: This class extends the BFO class \texttt{object aggregate}. Its subclasses include \texttt{Deck}, \texttt{Packet}, and \texttt{Fan}. An individual of type \texttt{Deck} is a deck of cards; one of type \texttt{Packet} is a group of between two and fifteen cards; and one of type \texttt{Fan} is a card fan.
\item \texttt{Card Magic Symbol}: This class extends the IAO class \texttt{symbol}, and includes subclasses like \texttt{Name (Playing Card)}, one of whose individuals is \texttt{Ace of Clubs}.
\item \texttt{Playing Card Quality}: This extends the BFO class \texttt{quality}, and allows expression of whether a playing card is facing up or down; or whether a playing card is a club, heart, spade, or diamond.
\item \texttt{Card Trick Method Instruction Entity}: This extends the IAO class \texttt{Information Content Entity}, and allows us to express declarative statements like, ``Perform Card Action X on the card at location Y.'' We will explore this class in detail in Section~\ref{section:concrete-example}.
\end{enumerate}
Each class in the Ontology of Card Sleights is axiomatized in $\mathcal{SROIQ}$(D) Description Logic, which is the strongest (most expressive) logic available in OWL DL. See section 3 of~\cite{description-logic-primer} for a definition and discussion of $\mathcal{SROIQ}$ Description Logic. We limit ourselves in this paper to stating that it is a first-order logic with extremely limited universal quantification, limited existential quantification, and additional requirements to ensure that questions about truth value always terminate, such as a requirement that properties not be cyclical. In principle, it is possible to push those limits. For example,~\cite{description-logic-rules} presents the technique of \emph{rolification} to create the logical implications of some universal quantification. However, we ended up removing all rolification axioms from the Ontology of Card Sleights, because they were causing automated reasoning programs either to hang or to run very slowly. (The reasoners that terminated said that our ontology was consistent, so we don't believe there was an error in our axiomatization). It may be possible in future to strengthen the axioms in our ontology without leaving OWL DL, as reasoning software advances.

We have used three automated reasoners to evaluate the Ontology of Card Sleights: FaCT++ 1.65~\cite{Tsarkov06fact++description}, HermiT 1.3.8~\cite{hermit-reasoner} and Pellet~\cite{pellet-reasoner}. All three evaluate our ontology as consistent and coherent. (An ontology is consistent if it has a model, i.e., if it implies no contradictory statements. An ontology is coherent if it has a model in which no class needs to be empty, i.e., every class can be populated by individuals.)
\subsection{From Conjuring Archive to Ontology of Card Sleights} \label{section:ocs}
We preserved most of the sleight classifications in the Conjuring Archive within our own ontology. However, the Conjuring Archive employs a naive tagging system, so there were some classifications we had to rebuild completely. As an illustrative example, consider the \emph{crimp}. Perhaps the simplest version of a crimp is the bent corner of a playing card. The magician (or gambler) can hand out a deck with a crimped card, and after the deck is cut, can locate cards of interest by finding the crimp by feel, and cutting to the crimped card. The word ``crimp'' is both a noun and a verb, there are many different kinds of crimps, and many ways in which a crimped card can be used.

The Conjuring Archive adds any trick or technique that uses a crimp to an overall crimp ``bucket.'' By contrast, the Ontology of Card Sleights contains both \texttt{Crimp} and \texttt{Crimp (use of)} classes. The class \texttt{Crimp (use of)} is a \texttt{Card Action}, and contains as subclasses the creation, removal and use of a crimp. The class \texttt{Crimp} is a \texttt{quality}, and it is a \texttt{quality} of a \texttt{Playing Card fiat object part} like a \texttt{Corner (Playing Card)}. BFO defines a \texttt{fiat object part} as a part of an object defined by human fiat, instead of natural division, such as a border between two countries. Where exactly the corner of a playing card starts and stops is a human decision, so a corner crimp is a quality of a fiat object part. Finally, the role a crimped card plays may change over time. In the example above, that of finding cards after a cut, the crimped card played the role of \texttt{Locator Card}. It might play a different role later in the magic show.

See Class Definition~\ref{class:crimp} for a full definition of the class \texttt{Crimp}. The axioms for the class are expressed in Manchester Syntax, a user-friendly compact syntax for OWL ontologies~\cite{manchester-syntax}. The property \texttt{has dedicated Card Action} is a sub-property of the BFO property \texttt{has participant}, which asserts that a \texttt{continuant} (roughly, BFO-ese for ``thing'') participates in a process. The formal definition of \texttt{has dedicated Card Action} also requires that the action be dedicated to the creation, destruction, or use of, the thing it is dedicated to. This definition is visible in the metadata of the relation, and we partially enforce it in OWL DL by declaring the relation to be \emph{inverse functional}: for any $x$, if $x$(\texttt{has dedicated Card Action})(\textbf{Crimp (use of)}), then $x$ must be a \textbf{Crimp}. Moreover, the companion axiom for \texttt{Crimp (use of)} is: [\texttt{Crimp (use of)}](\texttt{dedicated card action for})[\emph{only} \texttt{Crimp}]. The keyword \emph{only} is a restricted existential quantifier, that only ranges over individuals of type \texttt{Crimp}.

Class Definition~\ref{class:jordan-count} shows the definition of a representative Card Sleight, the Jordan Count. (The Jordan Count is a sleight in which the performer counts cards from one hand to the other, but the cards are not shown as they would be in a normal count.) The metadata fields link to two online resources: the Conjuring Archive, and the Conjuring Credits project~\cite{conjuring-credits}. Conjuring Credits is a volunteer effort to document the history and origins of theatrical magic. (For disclosure, we have contributed some material to Conjuring Credits.) We envision a user interface in which the Jordan Count has a ``profile page,'' and the user can click the links in the metadata to seamlessly move to related online content of projects that have already existed for years. Also note that the individual ``original Jordan Count'' has both a starting state and an ending state: four cards start in a particular order, and they end in another order. We do not consider sleights more granularly than that---while we could in principle model each time one hand takes a card from the other hand, instead we consider sleights as a black box, with a starting state and an ending state that is arrived at without intervening steps.
\begin{axiomatization}
\hfill \break
\textbf{Subclass of}
\begin{itemize}
\item[] \texttt{quality} (from BFO)
\end{itemize}
\textbf{Metadata}
\begin{itemize}
\item[] \textbf{definition} To bend one or a number of cards, so that they may be distinguished or located.
\item[] \textbf{Definition source} Erdnase, Expert at the Card Table, p. 28.
\item[] \textbf{Elucidation} This class categorizes physically crimped objects -- a crimped corner, a bridged deck, etc. For the process of putting a crimp in a card, see Crimp (use of).
\end{itemize}
\textbf{Axioms}
\begin{itemize}
\item[] \texttt{quality of} \emph{some} (\texttt{Playing Card} \emph{or} \texttt{Playing Card fiat object part} \emph{or} \texttt{Playing Card Object Aggregate})
\item[] \texttt{has dedicated card action} \emph{some} \texttt{Crimp (use of)}
\end{itemize}
\caption{Crimp} \label{class:crimp}
\end{axiomatization}
\begin{axiomatization}
\hfill \break
\textbf{Subclass of}
\begin{itemize}
\item[] \texttt{process} (from BFO) $\rightarrow$ \texttt{Card Action} $\rightarrow$ \texttt{Card Sleight} $\rightarrow$ \texttt{False Count}
\end{itemize}
\textbf{Metadata}
\begin{itemize}
\item[] \textbf{Definition} A false count (or variant of it) created by Charles Jordan.
\item[] \textbf{Conjuring Archive Link} \url{https://www.conjuringarchive.com/list/category/956}
\item[] \textbf{Conjuring Credits Link} \url{http://www.conjuringcredits.com/doku.php?id=cards:jordan_count}
\end{itemize}
\textbf{Individual: Jordan Count (4 as 4)}
\begin{itemize}
\item[] \textbf{Definition} The original Jordan Count.
\item[] \textbf{Axiom} \texttt{has participant} \emph{exactly} 4 \texttt{Playing Card}.
\item[] \textbf{Object Property} \texttt{has starting state} [\texttt{Jordan Count (4x4) starting state: Top card}]
\item[] \textbf{Object Property} \texttt{has ending state} [\texttt{Jordan Count (4x4) ending state: Top card}]
\end{itemize}

\caption{Jordan Count} \label{class:jordan-count}
\end{axiomatization}
\subsection{List of instructions} \label{section:list}
The Web Ontology Language is built on top of RDF syntax~\cite{rdf-syntax}. RDF, short for Resource Description Format, is often used in knowledge management applications. RDF allows declarations about the world with subject-relation-object triples, and includes some predefined vocabulary words, such as \texttt{rdf:List}. However, OWL uses RDF primitives under the hood to build OWL DL, and OWL DL does not recognize as lists the objects marked with \texttt{rdf:List}. We need to define our own list structure, in order to axiomatize and reason about the list of instructions that makes up the method of a card trick. The list structure we use is closely related to \texttt{OWLList}, the main class in the LIST ontology~\cite{owl-in-order}. This is the structure we use for the starting and ending state of sleights, introduced in Section~\ref{section:ocs}. It is also the structure we use for card trick instructions, which we will see now, and in Section~\ref{section:concrete-example} in greater detail.

We actually define two classes of lists. They are isomorphic, and we could use a single list structure, but the redundancy of definition provides us greater type safety. The two types of lists are \texttt{Card Action Instruction} and \texttt{Card Stack State Instruction}. The purpose of a \texttt{Card Action Instruction} is to state, ``Perform Card Action $X$ at location $Y$''; the purpose of a \texttt{Card Stack State Instruction} is to define the starting state of the card trick---the way in which the deck is stacked before anything happens. For both types of list, an individual \texttt{Card Trick Method} has a ``start here'' relation, which defines the head of the list---the first instruction in the method, the top card of the stack. Each node in the list points to another node of the same type, and OWL DL recognizes that the list has terminated because the final node of the list is an empty node, with no contents and no next node.

See Class Definition~\ref{class:card-action-instruction} for a formal definition of \texttt{Card Action Instruction}. There is a transitive relation \texttt{precedes card action instruction} which provides a linear order on the instructions. (A subrelation \texttt{immediately precedes card action instruction} is defined in the natural way, and plays a role in Section~\ref{section:concrete-example}.) Each individual Card Action Instruction directs the performance of one Card Action (the object of \texttt{perform card action}), upon the card at location referred to by the object of \texttt{focus on card location}.

The range of \texttt{focus on card location} is \texttt{Card Trick Location Instruction}, which in principle could refer to almost anywhere, such as ``in your pocket,'' or ``in the spectator's hands.'' At this point, though, we have only implemented references to a location in a stack of playing cards, like ``top card,'' ``fifth card,'' ``bottom card.'' This is achieved by using individuals with clever names like \texttt{Packet Location Focus: Top card of Packet}. Those individuals refer to the location, and the Card Stack State Instruction of the card trick, or the starting state of the sleight, determine which specific card is at that location.

Each list structure ends with an empty list node. An empty node precedes no other node, and contains nothing. Formally, the class \texttt{Empty Card Action Instruction} is equivalent to the class \texttt{Card Action Instruction} with the following additional axiom:
\begin{itemize}
\item[] \emph{not} (\texttt{precedes card action instruction} \emph{some} \texttt{owl:Thing}) \emph{and}
\item[] \emph{not} (\texttt{perform card action} \emph{some} \texttt{owl:Thing}) \emph{and}
\item[] \emph{not} (\texttt{focus on card location} \emph{some} \texttt{owl:Thing})
\end{itemize}
The class \texttt{owl:Thing} is an OWL primitive that contains every class. We now have enough machinery to transcribe a card trick.
\begin{axiomatization}
\hfill \break
\textbf{Subclass of}
\begin{itemize}
\item[] \texttt{information content entity} (from IAO) $\rightarrow$ \texttt{Card Trick Method Instruction Entity}
\end{itemize}
\textbf{Metadata}
\begin{enumerate}
\item[] \textbf{Definition} An instruction to perform (and/or how to perform) a Card Action.
\item[] \textbf{Example of usage} (1) Shuffle the deck. (2) Turn the top card face up. (3) Replace the palmed card onto the bottom of the deck.
\end{enumerate}
\textbf{Axioms}
\begin{itemize}
\item \texttt{precedes card action instruction} \emph{only} \texttt{Card Action Instruction}
\item \texttt{perform Card Action} \emph{only} \texttt{Card Action}
\item \texttt{focus on card location} \emph{only} \texttt{Card Trick Location Instruction}
\end{itemize}
\caption{Card Action Instruction} \label{class:card-action-instruction}
\end{axiomatization}
%
%
\section{A Concrete Example} \label{section:concrete-example}
We transcribe a card trick that uses the four Aces: Overture, created by Goldstein~\cite{focus}. The (simplified) effect of the trick is: the red Aces start out face up, and the black Aces start out face down; then, as if by magic, the black Aces are face up and the red Aces face down. That was a description of the effect, but for purposes of this paper, we focus on transcribing the method. Algorithm~\ref{algorithm:overture} provides pseudocode for the steps taken in the method of Overture.

Our first step in translating the Overture pseudocode into OWL DL, is to define a \texttt{Card Trick Method} individual that says, ``Begin with the defined starting state, and execute the first instruction.'' The class definition, and an individual encoding Overture, appear in Class Definition~\ref{class:card-trick-method}. We apply the list machinery from Section~\ref{section:list}. The starting state is encoded by a list structure of names of playing cards; the list's  first node is the individual \texttt{Overture starting state: top card}. The pseudocode of Algorithm~\ref{algorithm:overture} is encoded by a list structure of Card Actions, whose first node is the individual \texttt{Overture: first action}.

The actual lists that encode the starting state and pseudocode require lots of text to make small changes from one step to the next. This is a weakness of OWL DL that may be unavoidable. To implement this for the general public, we envision a Javascript portal that has access to optimized Javascript list libraries. Pure OWL DL would be used only to confirm coherency and consistency before pushing updates. We will not provide the transcription of Overture in its entirety, as it would take several pages, but we encourage the interested reader to download the Ontology of Card Sleights and search for ``Overture Method'' to view the transcription in its entirety.
\begin{algorithm}
\caption{Overture, by P. Goldstein} \label{algorithm:overture}
\textbf{Cards available at start}: the four Aces. \\
\textbf{Starting state}: All cards face up in a packet. From top down: Black Ace, Red Ace, Red Ace, Black Ace.\\
\begin{algorithmic}[1]
\State Spread in the hands (use of)[Packet]
\State Turn over (single card)[First Black Ace]
\State Turn over (single card)[Second Black Ace]
\State Close the spread
\State Jordan Count (4 as 4)
\State Through-the-Fist Flourish
\State Elmsley Count (4 as 4)
\State Spread in the hands (use of)[Packet]
\end{algorithmic}
\end{algorithm}
\begin{axiomatization}
\hfill \break
\textbf{Subclass of}
\begin{itemize}
\item[] \texttt{information content entity} (from IAO) $\rightarrow$ \texttt{Card Trick Method Instruction Entity}
\end{itemize}
\textbf{Metadata}
\begin{itemize}
\item[] \textbf{Definition} Instructions for the physical acts required to perform a card trick. Does not include script or theatrical blocking.
\end{itemize}
\textbf{Axioms}
\begin{itemize}
\item[] \texttt{is about} \emph{some} (\texttt{Playing Card} \emph{or} \texttt{Playing Card Object Aggregate})
\item[] \texttt{available cards} \emph{some} \texttt{Name (cards)}
\item[] \texttt{unavailable cards} \emph{some} \texttt{Name (cards)}
\item[] \texttt{has starting state} \emph{some} \texttt{Card Stack State Instruction}
\item[] \texttt{first card action instruction} \emph{only} \texttt{Card Action Instruction}
\end{itemize}
\textbf{Individual: Overture Method}
\begin{itemize}
\item[] \textbf{Object property:} \texttt{available cards}[\texttt{Ace of Diamonds}]
\item[] \textbf{Object property:} \texttt{available cards}[\texttt{Ace of Spades}]
\item[] \textbf{Object property:} \texttt{available cards}[\texttt{Ace of Clubs}]
\item[] \textbf{Object property:} \texttt{available cards}[\texttt{Ace of Hearts}]
\item[] \textbf{Object property:} \texttt{has starting state}[\texttt{Overture starting state: Top card}]
\item[] \textbf{Object property:} \texttt{first Card Action Instruction}[\texttt{Overture: first action}]
\end{itemize}
\caption{Card Trick Method} \label{class:card-trick-method}
\end{axiomatization}
\section{Magic ethics and ontology best practices}
\subsection{Magic ethics}
Magic \emph{performances} are protected by copyright in the USA, and they may have protection under international treaties because of limited coverage provided to choreography~\cite{own-it}. However, \emph{methods} for magic tricks---and, in our case, methods for card tricks---are not protected, as a general rule~\cite{selling-secrets}. Therefore, the ``magic economy'' attempts to be self-policing, where vendors are shunned if they sell secrets they did not themselves create~\cite{secrets-revealed}. The ubiquity of content piracy and ``card trick reveal'' videos on YouTube mean that the self-policing is not gong well, but we see no need to add to this problem ourselves.

We envision the Ontology of Card Sleights as a tool that remains under the hood in database searches. The taxonomy of the ontology will be public, but the individuals of the ontology will not. Our plan is to pair the ontology with a fully public ontology of magic trick plots and effects, so it is possible to answer search queries like, ``Magic tricks using at most six cards, with plot $P$, that use no more than two sleights.''
\subsection{Ontology best practices} \label{section:best-practices}
Chapter 3 of~\cite{arp2015building} recommends eight best practices for building domain ontologies with BFO. Our work satisfies all eight. We discuss each criterion briefly. The italicized words are (abbreviated) definitions from~\cite{arp2015building}; the plain text is our commentary about the Ontology of Card Sleights.
\begin{enumerate}
\item \textbf{Realism:} \emph{The ontology describes reality.} The classes of our movement writing are based on over 1500 publications, which appeared over the last 100 years. (These are not explicitly cited, but the metadata of most Card Actions includes links to Conjuring Archive pages of primary sources.) We are able to transcribe methods of card tricks published by professional magicians.
\item \textbf{Perspectivalism:} \emph{The ontology models reality from a modular perspective.} We only model methods of card tricks that do not require three-dimensional information, and our interest is in transcribing conceptual keyframes of motion. An ontology of drama might model the same data completely differently, using criteria of the trick's effect on an audience; or it might add performance data to allow three-dimensional theatrical blocking information.
\item \textbf{Fallibilism:} \emph{Must be revisable in the face of new discoveries.} The current version of the ontology is available via a permanent url service provided by the Internet Archive. That url links to a public github repository that contains both the ontology and a changelog, so errors and updates to the class structure are public. No descendent of Card Action is more than distance 4 from Card Action, so the vast majority of classes are orthogonal to one another, and will not be affected by the alteration or removal of another Card Action.
\item \textbf{Adequatism:} \emph{The entities of the domain are not reducible to other kinds of entities.} We take Card Action as a primitive.
\item \textbf{Reuse:} \emph{Existing ontologies should be reused whenever possible when building new domain ontologies.} The Ontology of Card Sleights imports both BFO and IAO. Also, it explicitly provides room to grow. For example: a card case is a prop within \texttt{Card Magic Object}, which itself is a subclass of \texttt{Theatrical Magic Object}. One could place this ontology inside a larger ontology of magic.
\item \textbf{Balance Utility and Realism:} \emph{Making unrealistic assumptions about the model in order to achieve short-term success, may harm the ontology's long-term utility.} We wrote formal definitions of every Card Action, based on personal review of at least three sources for each Card Action, and also read the Conjuring Archive notes on every source in their list. Therefore, we believe that the ontology classes will remain largely unchanged as we populate the database with individuals.
\item \textbf{Open-ended Design Process:} \emph{Ease in design, maintenance and updating.} We built the ontology with Protege, a popular, well-understood tool. The primary repository of sources we have used, the Conjuring Archive, has been online for over ten years. We are in the process of building a Javascript portal, which will provide the ability to update the ontology through a drag-and-click interface.
\item \textbf{Low-Hanging Fruit:} \emph{Begin with domain features that are easiest to define; work up to more complex or controversial features.} We started with magic trick methods, and we intend in the future to build an ontology of magic trick effects. There have been several attempts to categorize effects, and they are not all mutually consistent. There is far less debate over what constitutes a sleight like the Jordan Count, though there are sometimes disputes over proper credit.
\end{enumerate}
\subsection{Criteria for croudsourceability}
There has always been a significant obstacle to the adoption of Labanotation and other movement writing: to transcribe movement requires someone highly trained in transcription. Our ``Holy Grail'' objective would be a machine-readable movement writing with a user interface on a phone, such that a movement professional (\emph{not} an ontology professional) could transcribe a performance in real time. In other words, a user with movement domain knowledge would only need to use a phone for the same amount as needed to videorecord a performance with that same phone, and the app on the phone would provide the expert transcription ability. Therefore, in addition to the design criteria in Section~\ref{section:best-practices}, we employed design criteria to encourage crowdsourceability, i.e., to make it easier for users in the domain to correctly add individuals to the database. We built our movement ontology inside an upper ontology (BFO) that has also framed ontologies in other technical fields that are used by scientists, not ontologists. Beyond that, we used the following design criteria.

We define a $(5,5)$\emph{-substring-free set}: for any string $S$ of length 5 or greater, there are at most 5 members of the set that contain $S$. (We do not count a handful of ancillary phrases, like ``(use of)'', which we also do not treat as positive results in searches.) The practical motivation for this is that autofill of a search bar will produce at most five candidates, for any keyword. We doubt we can maintain this property over all possible database individuals, but we have ensured that the class names of Card Actions are $(5,5)$-substring-free. That way, an app interface can provide a gesture-distance of at most five keystrokes and one button click, in order to reach the list of database individuals of interest.

We achieve $(5,5)$-substring-free-ness in part because we allow classes to have a primary name and an alternative name. Primary names must all be distinct, while alternative names need not be, and an interface might need to disambiguate them through a list of radio buttons (gesture-distance 1). As an example, there is a sleight with a somewhat controversial history, due to acrimonious debates over who created it. Currently, the primary name for this class of sleights is ``Convincing Control,'' and the alternative term is ``Immediate Bottom Placement,'' because the written record currently favors the person who named the sleight Convincing Control as the original creator of the sleight. If new data surfaces, we can change the priority of the names without affecting database search results.
\section{Conclusion}
We have presented a system of movement writing for methods of card tricks. While we believe our work to be an advance in the scientific understanding of human motion, it may be that the most relevant extension of our work will be into the motion of robots, not humans. Our approach succeeded, in part, because of the algorithmic nature of card tricks. Much human motion is less algorithmic. Robotic motions, by contrast, tend to have clearly-defined starting states, objectives, and transition rules. We believe that might be a fruitful area for future work.
\subsubsection*{Acknowledgements}
We would like to thank Denis Behr, Charlie Faraday, Carisa Hendrix and Stephen Minch for helpful discussions.
\bibliographystyle{IEEEtran}
\bibliography{OntologyBibliography}

\end{document}